\documentclass[runningheads]{llncs}
\usepackage{graphicx}
\usepackage{tabularx}
\newcolumntype{C}{>{\centering\arraybackslash}X}
\newcolumntype{R}{>{\raggedleft\arraybackslash}X}
\newcolumntype{L}{>{\raggedright\arraybackslash}X}
\newcolumntype{P}[1]{>{\raggedleft\arraybackslash}p{#1}}
\usepackage{xcolor}
\usepackage{hyperref}
\usepackage{enumitem}

\usepackage{subcaption}

\begin{document}
\title{Leveraging Data Augmentation for Process Information Extraction}
%
%
\author{Julian Neuberger\inst{1}\orcidID{0009-0008-4244-7659} \and
Leonie Doll\inst{1} \and
Benedikt Engelmann\inst{1} \and
Lars Ackermann\inst{1}\orcidID{0000-0002-6785-8998} \and
Stefan Jablonski\inst{1}}
\authorrunning{J. Neuberger et al.}
%
\institute{University of Bayreuth, Bayreuth, Germany
\email{firstname.lastname@uni-bayreuth.de}}
\maketitle              
\begin{abstract}
Business Process Modeling projects often require formal process models as a 
central component. High costs associated with the creation of such formal 
process models motivated many different fields of research aimed at automated 
generation of process models from readily available data. These include process
mining on event logs and generating business process models from natural
language texts. Research in the latter field is regularly faced with the 
problem of limited data availability, hindering both evaluation and development
of new techniques, especially learning-based ones.

To overcome this data scarcity issue, in this paper we investigate the 
application of data augmentation for natural language text data. Data 
augmentation methods are well established in machine learning for creating
new, synthetic data without human assistance. We find that many of these 
methods are applicable to the task of business process information extraction, 
improving the accuracy of extraction. Our study shows, that data augmentation 
is an important component in enabling machine learning methods for the task of 
business process model generation from natural language text, where currently
mostly rule-based systems are still state of the art. Simple data augmentation 
techniques improved the $F_1$ score of mention extraction by 2.9 percentage 
points, and the $F_1$ of relation extraction by $4.5$. To better understand how 
data augmentation alters human annotated texts, we analyze the resulting text, 
visualizing and discussing the properties of augmented textual data.

We make all code and experiments results publicly available\footnote{Code for 
our framework can be found at \url{https://github.com/JulianNeuberger/pet-data-augmentation}, 
detailed results for our experiments as MySQL dump can be 
downloaded from \url{https://zenodo.org/doi/10.5281/zenodo.10941423}}.

\keywords{Business Process Extraction  \and Data Augmentation \and Natural 
Language Processing.}
\end{abstract}

\section{Introduction}\label{sec:intro}

It has been shown that a major share of time planned for Business Process 
Management (BPM) projects is spent on the acquisition of formal business
process models~\cite{friedrich2011process}. This fact motivated a whole host
of work done on automated generation of business process models from varying,
readily available sources, such as event logs, or natural language process 
descriptions. The latter area has seen increasing attention in recent 
years~\cite{friedrich2011process,van2019extracting,ferreira2017semi,%
quishpi2020extracting,ackermann2021data,neuberger2023beyond}. Most of these
approaches can be formulated as a two-step process, where first the 
business process relevant information is extracted from text, and then a 
concrete model is generated from this information.

Many approaches proposed for the task of business process information extraction (BPIE) 
from texts are 
still rule-based. This means they extract the information needed for building 
a formal business process model by applying rules defined by human experts.
These rules are usually optimized for a specific dataset, industry sector,
or language. For this reason, such systems usually achieve impressive
results for their intended, very limited, application domain, but are difficult 
to transfer to new ones. Alternative approaches like data driven approaches, 
often called machine learning (ML) methods, infer
rules directly from data, making them a lot easier to be adapted to
new datasets, domains, or languages. Ideally, adapting an ML approach
involves only training on new data. However, this need for data, 
both for the initial development, as well as for potential adaptations, 
is what typically impedes application of ML methods at first. Compared 
to other disciplines also using machine learning, datasets in BPM are 
relatively small, especially for the task of generating business process 
models from natural language text. Data sets for this task are expensive 
to generate, as time-consuming manual annotation by experts is required, 
in which raw text data is enriched with the desired results (e.g. process 
entities) in order to provide the ML methods with a basis for learning.
Previous work tried to solve this issue by leveraging out-of-domain data, 
e.g., via pretrained word embeddings~\cite{ackermann2021data}, or less 
expressive models, which are easier to train~\cite{neuberger2023beyond} on
small datasets. Other fields of related research, such as computer vision, natural 
language processing (NLP), or audio analysis, tackle the issue by synthesizing 
new data samples from the existing dataset. This concept is called data 
augmentation (DA), injecting controlled perturbations. These perturbations
can be structural changes (e.g., rotations), as well as added noise.

\begin{figure}[b]
    \centering
    \includegraphics[width=\textwidth]{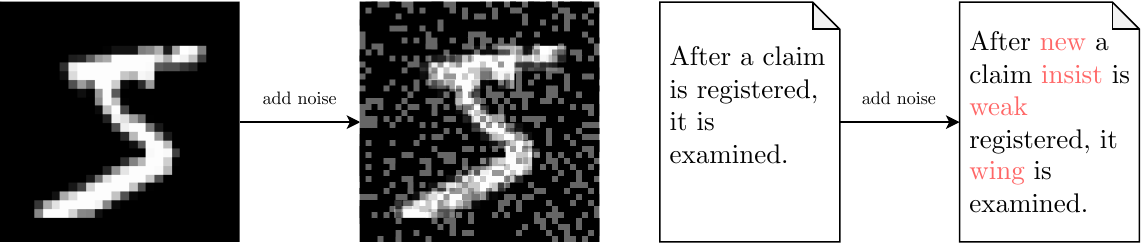}
    \caption{Example for adding noise to an image (left) and to a sentence (right).
    The image keeps its semantics, while the sentence looses it.}
    \label{fig:computer-vision-vs-nlp}
\end{figure}

Data augmentation has already been shown to be useful in a BPM context, 
such as for the task of next activity prediction, where it is proposed 
as a solution for the problem of rare process executions, as well as for 
extrapolating gradual changes in the way processes are executed
\cite{kappel2021leveraging,kappel2021evaluating}. The authors show that 
simple data augmentation strategies 
like swapping, inserting, or deleting events in the record of a process execution 
can result in significant accuracy gains~\cite{kappel2021evaluating}. Encouraged
by those results, we want to analyze how data augmentation can be applied to
improve existing methods for generating process models from text. Due 
to the nature of natural language, introducing noise without changing the semantics
of a data sample is much harder, compared to, for example, computer vision tasks. See Figure
\ref{fig:computer-vision-vs-nlp} for an example, where introducing noise into an
image still keeps its semantics (image of a handwritten ``5'') intact. Introducing
noise in the form of random words into a single sentence on the other hand, can
change its semantics significantly, to a point, where it is hard to understand even for humans. We will discuss this fact in more detail in Section
\ref{sec:background} and give examples for data augmentation techniques, which
preserve semantics.

To further structure our understanding of applying data augmentation in the 
context of generating business process models from natural language text, we 
pose three research questions. 

\begin{enumerate}[itemindent=.7cm, label=\textbf{RQ\arabic*}]
    \item\label{rq:simple-augmentation} Can simple data augmentation 
    techniques, including swapping, deleting, or randomly inserting words 
    into sentences increase the performance of machine learning methods 
    for BPIE, measured as the harmonic
    mean of precision and recall?
    \item\label{rq:complex-augmentation} Does the use of large language models
    in data augmentation, such as for so called \emph{Back Translation}
    techniques, provide a significant advantage over simpler, rule-based
    methods?
    \item\label{rq:characteristics} What characteristics of the natural 
    language text data are changed by augmentations, and how do they 
    affect different extraction tasks?
\end{enumerate}

The rest of this paper is structured in the following way: Section~\ref{sec:background}
describes the background of data augmentation in a NLP environment. In Section 
\ref{sec:related-work} we give an overview of work related to our study. Section
\ref{sec:experiment-setup} defines the setup for our experiments. We then present our
results in Section~\ref{sec:results}. Finally, in Section~\ref{sec:conclusion} we 
discuss limitations, describe future work, and draw a conclusion.

\section{Background}\label{sec:background}

\begin{figure}
    \centering
    \includegraphics[width=\textwidth]{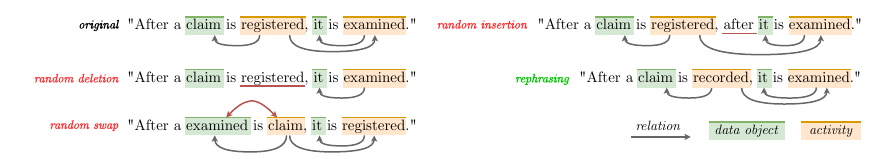}
    \caption{Examples for four different data augmentation techniques.
    \emph{Random deletion}, \emph{random swap}, \emph{random insertion} (all 
    written in red), are all not preserving the semantics of a sample and its 
    label. \emph{Rephrasing} (green) is an example for a technique that does.}
    \label{fig:augmentation-example}
\end{figure}

\textbf{Data augmentation} describes a suite of techniques
originally popularized in computer vision~\cite{shorten2021text},
where simple operations, such as cropping, rotating, or introducing
noise into images greatly improved performance of machine learning
algorithms used for classification of images. These operations 
usually preserve the semantics of input data, meaning that an 
image containing an object will still depict the same object after 
its data have been augmented, for example, they have been overlaid 
with noise. This property is called \emph{invariance}~\cite{zoran2009scale}, 
and is harder to hold for NLP data \cite{feng2021survey}. An example 
for this fact is depicted in Figure~\ref{fig:augmentation-example}. 
Changing random tokens (e.g., words) in a sentence may alter semantics 
to a point, where relevant elements or relations between those elements 
are no longer present after augmentation. Additionally, annotations may 
be lost, if transformations are applied and afterwards these changes cannot 
be traced. This might happen, when, for example, an entire sentence is 
translated into another language and then is back-translated typically 
leading to a rephrased version of the original text. Since it is 
not clear, which parts of the new sample correspond to the original one, 
annotations of process-relevant elements do not apply to the new sample.
For this reason, research on data augmentation techniques has been conducted,
which are specifically designed for information extraction tasks in the 
NLP domain\cite{jiang2021cori,liu2021machine,erdengasileng2022pre}. 
These techniques use additional resources, such as pretrained large 
language models, to augment training samples, while keeping their semantics 
intact.

\begin{figure}
    \centering
    \includegraphics{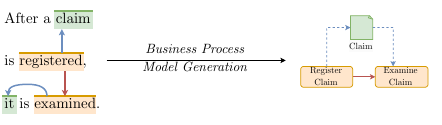}
    \caption{Example for a fragment of a natural language business 
    process description and its corresponding business process model 
    fragment in BPMN.}
    \label{fig:process-model-example}
\end{figure}

\textbf{Process Relevant Information Extraction} from natural language is a research 
field immediately relevant for information systems, as business process models 
are often a central part for process aware information systems. Discovering and 
creating these process models is an expensive task~\cite{friedrich2011process} 
and a lot of work has been done on extracting them from natural language text 
directly~\cite{friedrich2011process,van2019extracting,ferreira2017semi,%
ackermann2021data,bellan2022extracting}. These texts describe a business process 
in natural language as technical documentation, maintenance handbooks, or 
interview transcripts. Sequences of words (spans) in these texts contain 
information that is relevant to the business process, such as \emph{Actors}
(persons or departments involved in the process), \emph{Activities} (tasks
that are executed), or \emph{Data Objects} (physical or digital objects 
involved in the process). Extracting this information is therefore a sequence
tagging task, and can be framed as \textit{Mention Detection} (MD). 
Mentions relate to each other, e.g., defining the order of execution for 
two Activities, or which Actor executes the Activity. Predicting and classifying 
these relations is called 
\emph{Relation Extraction} (RE). Refer to Figure~\ref{fig:process-model-example}
for an example of this process. It shows a fragment of a larger description of a 
process from the insurance domain, where insurance claims have to be registered
in a system and subsequently examined by an employee. The spans \emph{claim} and
\emph{it} are annotated as Data Objects (the claim in question, in green).
Activities executed by a process participant are marked in orange. These four
spans can now be transformed into business process model elements for a target
notation language (here BPMN\footnote{See specification at \url{https://www.bpmn.org/}.}). 
How these elements
interact with each other can also be extracted from the text fragment, e.g.,
the \emph{Flow} of activity execution between the mentions \emph{registered}
and \emph{examined}, depicted as an orange arrow.

Developing approaches towards automated extraction of process relevant information 
requires data to test performance, and train models, if applicable. The currently
largest collection of human-annotated process descriptions is called PET
\cite{bellan2023pet}. It contains 45 natural language process descriptions, and 
is annotated with 7 types of process relevant entities (e.g., Actors, Activities, 
Data Objects), as well as 6 types of relations between them (e.g., Flow between 
Activities). In total the dataset contains less than 2,000 examples for both 
relations and entity mentions. For comparison, typical datasets for related tasks,
like Knowledge Graph completion contain more than 200 times as many. For example, 
the popular \emph{FB15k} dataset comprises more than 500,000 relation examples
\cite{bordes2013translating}. Datasets for extraction of named entities and their 
relations have similar extents, e.g., the DocRed dataset, which contains more than
1,500,000 relation examples~\cite{yao2019docred}. This fact makes PET a prime 
candidate for data augmentation techniques, in order to make the most out of 
the limited amount of training examples. We show this in our experiments using PET
for the tasks MD and RE in BPIE. To our knowledge our work is the first 
to attempt applying NLP data augmentation to the BPIE task. 

\section{Related Work}\label{sec:related-work}

Data augmentation techniques applied in this paper are largely based on the 
ones available in the \emph{NL-Augmenter} framework~\cite{dhole2021nl}.
NL-Augmenter provides a list of more than 100 data augmentation 
techniques, which are 
suitable for varying tasks like text classification, sentiment 
analysis, and even tagging. We discuss how we adapted these 
transformations to the PET data format in more detail in Section
\ref{sec:experiment-setup}. Not all transformations are relevant for
this work, and we have to exclude most of them, as they are not fitting
for BPIE. Details of our exclusion criteria can be found in Section 
\ref{sec:experiment-setup}.

In~\cite{kappel2023model} the authors evaluate nine simple data 
augmentation techniques (e.g., random deletion) on a total of seven
event logs, using seven different models. Our paper follows a similar line 
of thought for BPIE, instead of predictive process monitoring. The
transformations we employ differ significantly from theirs in two core 
aspects. First, transformations used in this paper are more complex, 
owing to the more complex character of natural language. While their 
work focused on reordering events in a log of a process execution,
our work uses transformations that are concerned with replacing, 
extending, or modifying sequences of text, while preserving any 
annotations present in the data. Second, transformations used
in our work often require external resources. These resources can
be explicit, i.e., databases like WordNet~\cite{miller1995wordnet}, 
which contains lexical information such as synonyms, antonyms, or 
hypernyms of words. They can also be implicit, such as large language
models, which contain knowledge about natural language, obtained by
unsupervised training on huge amounts of textual data~\cite{devlin2018bert}.

The techniques we present in our paper mainly benefit work that already
exists in the field of BPIE. Therefore, approaches towards BPIE based on 
machine learning are related to this work. These approaches can be 
separated into two main fields of research. \textit{(1)} learning approaches,
which use the data to train a machine learning models, e.g., a neural 
network~\cite{ackermann2021data}, conditional random fields~\cite{bellan2023pet}, 
or decision trees~\cite{neuberger2023beyond}. \textit{(2)} prompting based
approaches that use the data for engineering input for large language 
models (e.g., GPT)~\cite{kampik2023large,klievtsova2023conversational}, 
or use the data for so called \textit{in context learning}, by providing 
examples in the input itself~\cite{bellan2022extracting}.

Automated extraction of information relevant to business processes from 
natural language text descriptions can be seen as a special case of automated
knowledge graph construction or completion~\cite{DBLP:conf/aiia/BellanDG22}.
We therefore consider techniques for automated knowledge graph construction
and completion as distantly related work, which could still benefit from
the augmentation techniques we analyze in this paper. Nonetheless, we focus on
methods of BPIE in this paper, as potential solution for this field's small datasets.

\section{Experiment Setup}\label{sec:experiment-setup}

The NL Augmenter framework provides a total of 119 data augmentation 
techniques, but not all of them are applicable to the task at hand. We 
therefore define four criteria for exclusion: 
\textbf{(EC1)} The technique does not apply to the English 
language. \textbf{(EC2)} The technique alters the spelling of 
tokens. \textbf{(EC3)} The data augmentation 
technique does not work for supervised data, e.g., it corrupts target 
annotations. \textbf{(EC4)} The technique uses task-, 
and/or domain-specific resources, such as dictionaries, or databases,
which often do not exist for BPM data, and are hard to create given the
diversity of BPM application domains.
Applying these exclusion criteria results in 19 data augmentation techniques 
relevant for the task of business process information extraction from natural 
language text.

\subsection{Data Augmentation Effects}

The data augmentation techniques we selected synthesize samples with 
three core characteristics. 

\textbf{(1)}\label{effect:ling-var} Increased linguistic variability, i.e.,
augmented text uses a larger vocabulary to describe the same, or at least,
a similar business process\footnote{The augmentation technique might change
information in the text, which changes the process overall, e.g., by replacing
original actors with new, artificial ones.}. The most prominent examples for such
techniques are the \emph{Back-Translation} techniques. These
use a large language model, e.g., BERT~\cite{devlin2018bert} to translate
the process description to a different language and subsequently translate
it back to the original language -- here English. Since data augmentation 
techniques must not alter the annotations of entities, we only translated 
spans of text, not the entire document at once. Take for example, the running 
example \emph{After a claim is registered, it is examined.} Here four spans
are annotated as entities -- \emph{a claim}, \emph{registered}, 
\emph{examined}, and \emph{it}. Additionally there are three remaining spans,
that do not correspond to entities: \emph{After a}, \emph{is}, \emph{is}.
By back-translating these seven spans separately, we obtain variation in 
their wording (\emph{surface form}), but are still able to preserve 
annotations. Samples synthesized in this way are especially useful for
making methods for the MD task generalize better and more robust.

\textbf{(2)}\label{effect:span-length} Variations in span length. Many spans 
of a given entity type,
e.g., Actors are very uniform in length across examples. This is a result of
several factors, but most apparent actors are often identified by their job
title, e.g., \emph{the clerk}, or the department, e.g., \emph{the secretary
office}. These titles and departments are very short phrases, and longer ones
are abbreviated, reducing their length to two or less tokens, e.g., \emph{
the MPOO}. Even though their expanded form may not be known, expanding some
of these spans to suitable phrases, e.g., \emph{Manager, Post Office Operations}, 
creates samples with longer surface forms. This in turn, may improve the 
robustness of the MD extractors, as well as the generalization capabilities of 
RE methods.

\textbf{(3)}\label{effect:directionality} Directionality of relations between 
mentions. The order of 
appearance for mentions that form a relation, is very uniform in the current
version of PET. This is especially apparent, when looking at the base-line
extraction rules defined by the original authors of PET: Here the order of 
appearance of Activities and Actors is exploited, to form the \emph{Actor 
Performer} and \emph{Actor Recipient} relations~\cite{bellan2023pet}. These 
relations define the Actor, that performs an Activity, and the Actor, on which
an Activity is performed. The Actor left of an Activity is assigned the former,
while the Actor right of that Activity is assigned the latter. In this example
order uniformity can lead to less robust models, as they rely on this and 
subsequently make wrong predictions given different linguistic constructs.
Synthesizing samples with a different order may encourage models to consider
linguistic features (context) rather than just the order of mentions in a sentence
during prediction.

\subsection{Finding Optimal Configurations}

\begin{figure}[bt]
    \centering
    \includegraphics[width=\textwidth]{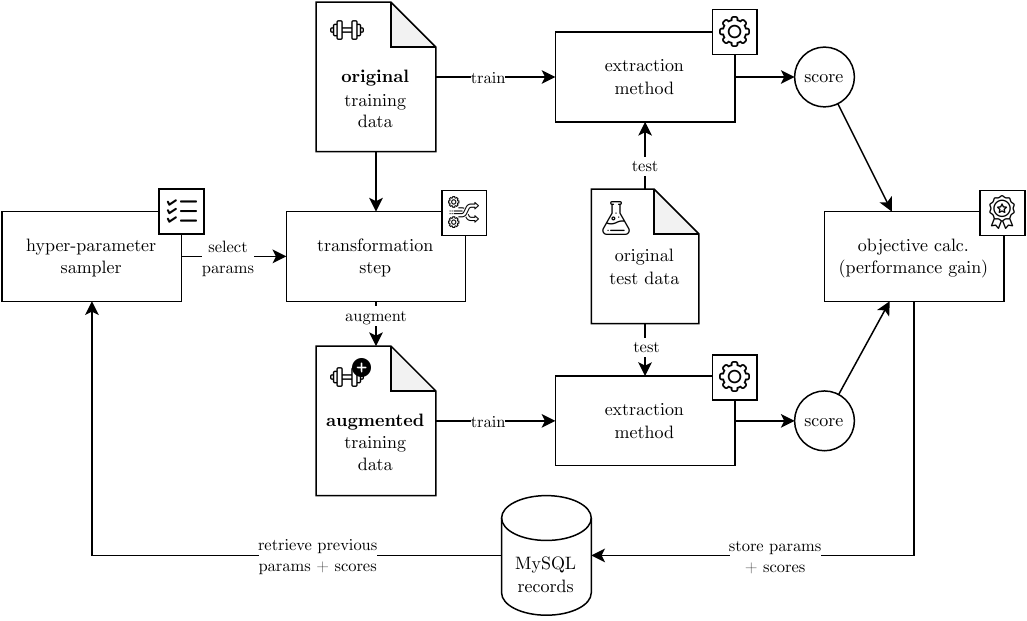}
    \caption{Choosing optimal configurations for data augmentation techniques.}
    \label{fig:hyper-param-optimization}
\end{figure}

Each of the data augmentation techniques we selected can potentially be adjusted by 
several parameters, which control how augmented samples are synthesized. 
A typical example for such a parameter is the number of inserted 
tokens. Increasing this number would result in a sample, which is more
perturbed compared to a sample where fewer tokens are inserted. We 
consider optimally choosing such parameters for a given technique a 
\emph{hyper-parameter optimization} problem. Hyper-parameter optimization is 
defined as finding a configuration of parameters so that a given 
objective (metric to optimize) is minimal or rather maximal, depending 
on the case. Here, we want to maximize the performance gain that the 
application of a data augmentation technique has. To that end we run a 
5-fold cross-validation of the extraction step (MD, RE) with the original,
unaugmented data. We then select a configuration for the given technique
and run the same 5-fold cross-validation, but augment the training data of 
each fold with the data augmentation technique. We define the difference 
between the scores of these two models on the (unaugmented) test dataset as 
the \emph{performance gain} and use it as maximization objective for our 
hyper-parameter optimization. Each data augmentation technique is optimized in 
25 runs (\emph{trials}) using Optuna~\cite{akiba2019optuna} and a Tree-Structured 
Parzen Estimator for selecting parameter values \cite{bergstra2011algorithms}. 
We depict this process in Figure \ref{fig:hyper-param-optimization}.

%
%
The most effective data augmentation techniques can then be used to supplement 
the existing PET dataset, as well as any future datasets used for BPIE. 
%
%
In the following Section~\ref{sec:results} we present the results of the 
experiment described here and discuss their implications.

\section{Results}\label{sec:results}

In this section we will discuss results for the experiments 
we defined in the previous section. Table~\ref{tab:results}
lists the differences of all data augmentation techniques compared 
to a run on un-augmented data. All differences are measured as the 
micro-averaged $F_1$ score. Concluding from our results, we find that the RE task can benefit significantly from most of the data augmentation techniques we selected and tested. 
Transformations that reorder tokens, like \emph{Shuffle Within Segments},
\emph{Sentence Reordering}, or \emph{English Mention Replacement} seem to be
less useful, compared to other techniques. This is most likely rooted in the 
change in directionality of relations (effect \textbf{(3)} from Section
\ref{sec:experiment-setup}). Since these transformations do not take any 
context into account, such changes may be breaking semantics of relations
in a sentence.

Transformations based on large language models, especially back translation 
techniques, like \emph{Multilingual Back Translation}, which translate a 
sentence fragment twice, are very time-intensive. Yet, improvements in 
relation extraction performance is not significant, when comparing them to
more lightweight approaches, e.g., \emph{Synonym Substitution}, which uses
WordNet to rephrase text sequences. In our experiments using these large 
language model based methods is not worth the increase in computing power
and time. While the MD task can still benefit from all data augmentation techniques,
it does so to a lesser extent when compared to the RE task. This indicates
a model, that is already more stable, and generalizes better. Transformations
that alter the amount of tokens in mentions, such as \emph{Random Word Deletion},
\emph{Synonym Insertion}, or \emph{Subsequence Substitution for Sequence 
Tagging}, result in lesser improvements, compared to paraphrasing methods, 
such as \emph{AntonymsSubstitute}, \emph{BackTranslation}, or \emph{Synonym
Substitution}. Similar to the RE task, the MD task does not benefit 
significantly more from resource and time intensive, large language model
based augmentation techniques for paraphrasing, compared to their simpler
counterparts.

\begin{table}[hbt]
\centering
\begin{tabularx}{\textwidth}{Lr|L|P{1.1cm}P{1.1cm}}
    \textbf{Technique} & \textbf{Id} & \textbf{Description} & \textbf{MD} & \textbf{RE} \\

    \hline

    Unaugmented & & results for un-augmented data & 0.695 & 0.759 \\
    
    \hline
    
    Adjectives Antonyms Switch & \href{https://github.com/GEM-benchmark/NL-Augmenter/tree/main/nlaugmenter/transformations/adjectives_antonyms_switch}{B.3} & use antonyms of adjectives & $+0.024$ & $+$\textbf{0.045} \\
    
    AntonymsSubstitute (Double Negation) & \href{https://github.com/GEM-benchmark/NL-Augmenter/tree/main/nlaugmenter/transformations/antonyms_substitute}{B.5} & substitute even number of words with antonyms & $+$\textbf{0.025} & $+0.042$ \\

    Auxiliary Negation Removal & \href{https://github.com/GEM-benchmark/NL-Augmenter/tree/main/nlaugmenter/transformations/auxiliary_negation_removal}{B.6} & remove negated auxiliaries & $+$\textbf{0.025} & $+0.039$ \\

    BackTranslation & \href{https://github.com/GEM-benchmark/NL-Augmenter/tree/main/nlaugmenter/transformations/auxiliary_negation_removal}{B.8} & translate to German, then back to English & $+$\textbf{0.025} & $+0.036$ \\

    Concatenate Two Random Sentences & \href{https://github.com/GEM-benchmark/NL-Augmenter/tree/main/nlaugmenter/transformations/concat_monolingual}{B.24} & remove punctuation between sentences & $+0.023$ & $+0.042$ \\

    Contextual Meaning Perturbation & \href{https://github.com/GEM-benchmark/NL-Augmenter/tree/main/nlaugmenter/transformations/contextual_meaning_perturbation}{B.26} & replace words with use of pretrained language model & $+0.004$ & $+0.040$ \\

    English Mention Replacement for NER & \href{https://github.com/GEM-benchmark/NL-Augmenter/tree/main/nlaugmenter/transformations/entity_mention_replacement_ner}{B.39} & replace mention with one of the same type in document & $+0.015$ & $+0.036$ \\

    Filler Word Augmentation & \href{https://github.com/GEM-benchmark/NL-Augmenter/tree/main/nlaugmenter/transformations/filler_word_augmentation}{B.40} & introduce ``uhm'', ``I think'', ... & $+0.021$ & $+$\textbf{0.043} \\

    Multilingual Back Translation & \href{https://github.com/GEM-benchmark/NL-Augmenter/tree/main/nlaugmenter/transformations/multilingual_back_translation}{B.62} & see B.8, language is parameter & $+0.022$ & $+0.041$ \\

    Random Word Deletion & \href{https://github.com/GEM-benchmark/NL-Augmenter/tree/main/nlaugmenter/transformations/multilingual_back_translation}{B.79} & delete random words & $+0.011$ & $+0.034$ \\

    Replace Abbreviations and Acronyms & \href{https://github.com/GEM-benchmark/NL-Augmenter/tree/main/nlaugmenter/transformations/replace_abbreviation_and_acronyms}{B.82} & replace acronyms with full length expression and v.v. & $+0.020$ & $+0.042$ \\

    Sentence Reordering & \href{https://github.com/GEM-benchmark/NL-Augmenter/tree/main/nlaugmenter/transformations/sentence_reordering}{B.88} & reorder sentences & $+0.024$ & $+0.034$ \\

    Shuffle Within Segments & \href{https://github.com/GEM-benchmark/NL-Augmenter/tree/main/nlaugmenter/transformations/shuffle_within_segments}{B.90} & shuffle tokens in mentions & $+0.021$ & $+0.041$ \\

    Synonym Insertion & \href{https://github.com/GEM-benchmark/NL-Augmenter/tree/main/nlaugmenter/transformations/synonym_insertion}{B.100} & insert synonym before word & $+0.019$ & $+$\textbf{0.043} \\

    Synonym Substitution & \href{https://github.com/GEM-benchmark/NL-Augmenter/tree/main/nlaugmenter/transformations/synonym_substitution}{B.101} & substitute word with synonym & $+0.023$ & $+0.040$ \\

    Subsequence Substitution for Sequence Tagging & \href{https://github.com/GEM-benchmark/NL-Augmenter/tree/main/nlaugmenter/transformations/tag_subsequence_substitution}{B.103} & replace sequence with another sequence with same POS tags & $+0.019$ & $+0.033$ \\

    Transformer Fill & \href{https://github.com/GEM-benchmark/NL-Augmenter/tree/main/nlaugmenter/transformations/transformer_fill}{B.106} & replace tokens using language model & $+0.022$ & $+0.041$ \\

    Random Insert &  & insert random tokens & $+0.020$ & $+$\textbf{0.043} \\

    Random Swap &  & swap position of tokens & $+$\textbf{0.029} & $+0.033$ \\
    \hline
\end{tabularx}
\vspace{.1cm}
\caption{Detailed results for all transformation steps, for both the MD and RE task.
Column \textit{Id} refers to the identifier defined in~\cite{dhole2021nl}. It links to
the source code for this technique. Top three results are set 
in \textbf{bold} face. All results are the averages of a 5-fold cross validation
on the entire dataset.}
\label{tab:results}
\end{table}

\begin{figure}[bt]
    \centering
    \includegraphics[width=\linewidth]{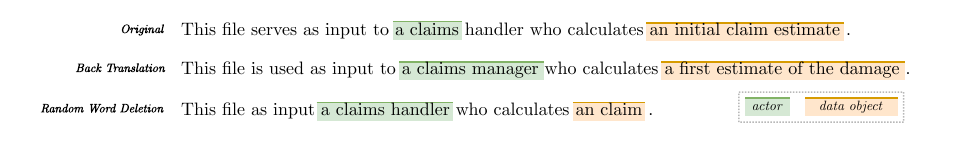}
    \caption{Two examples for the effects of data augmentation techniques.}
    \label{fig:augmentation-examples}
\end{figure}

Based on our observations, we can answer research question
\ref{rq:simple-augmentation}, with ``Yes''. Simple data augmentation techniques
like swapping tokens, deleting them, or inserting random tokens do have
a significant benefit. The RE task benefits more from them, than the MD
task does. Since nearly all augmentations have a net positive impact, we
argue that the perturbations act as controlled noise, very similar to
techniques used for training deep neural networks. There, models that are 
more robust and generalize better are created, simply by adding noise to 
the training data~\cite{srivastava2014dropout}.

The use of large language models in data augmentation techniques brings 
with it a significant increase in resources needed, both in terms of
computing time, hardware requirements, and power consumption. Based on 
our experiments, this is not worthwhile for improving the business 
process information extraction approaches we used. Back-translation
techniques, such as \textit{B.8}, \textit{B.62}, and especially 
\textit{B.26} do not provide benefits in the MD and RE tasks, that 
would warrant their additional needs in hardware (GPUs), and runtime, 
which was several orders of magnitude higher, compared to simpler 
augmentation techniques. We therefore have to answer 
\ref{rq:complex-augmentation} with ``No''.

\begin{figure}[bt]
\centering
\begin{subfigure}[t]{.50\textwidth}
    \centering
    \includegraphics[width=\linewidth]{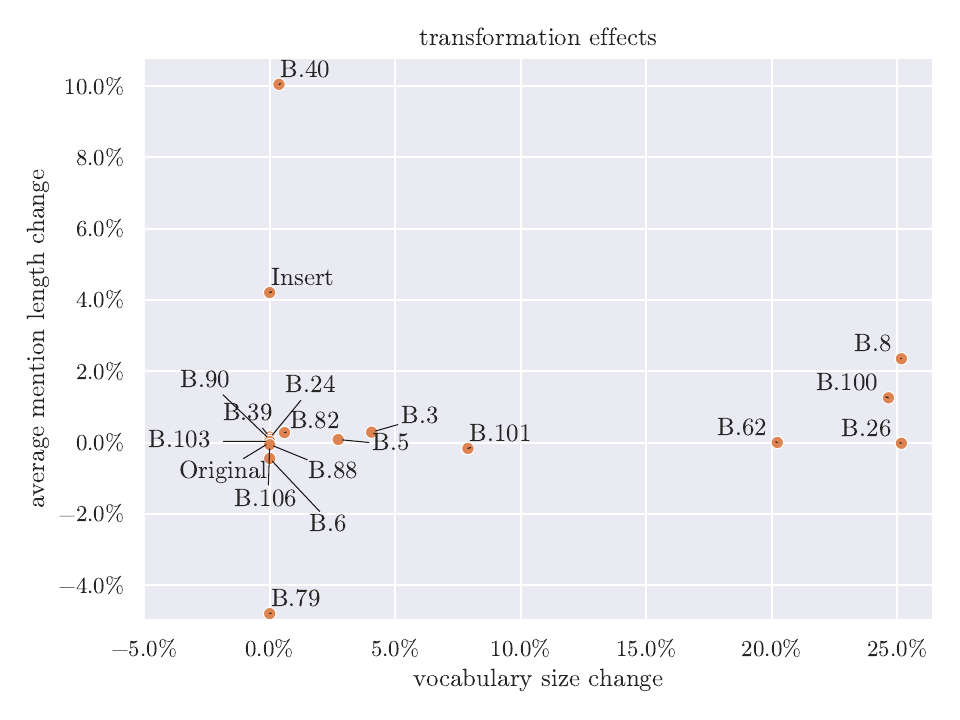}
    \caption{}
    \label{fig:trafo-effects}
\end{subfigure}%
\hfill%
\begin{subfigure}[t]{.50\textwidth}
    \centering
    \includegraphics[width=\linewidth]{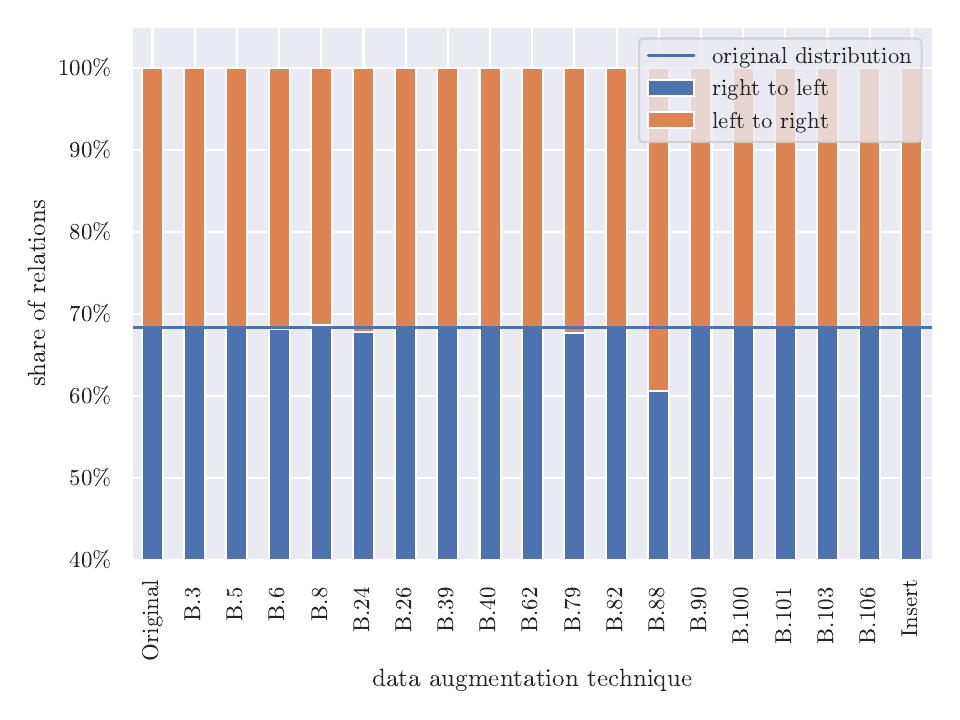}
    \caption{}
    \label{fig:directionlity}
\end{subfigure}%
\caption{\textit{(a)}: Effects on vocabulary size and the average length of 
mentions in tokens. \textit{(b)}: Effects of techniques on relation direction.  }
\label{fig:stochasticity_and_num_shots}
\end{figure}

To answer research question~\ref{rq:characteristics}, we defined three 
characteristics of textual data in Section~\ref{sec:experiment-setup} 
that are changed by the data augmentation techniques we selected. These
characteristics are visualized in Figures~\ref{fig:trafo-effects} and
\ref{fig:directionlity}. Figure~\ref{fig:trafo-effects} shows the 
``landscape'' of 
data augmentation techniques evaluated in this paper. Three groups 
of techniques emerge. The first one is a group of techniques that
only marginally increase the number of tokens in mentions, and keep
the size of the vocabulary roughly the same. These techniques mainly 
change the context (i.e., the text that does not contain immediately
process relevant information), or the structure of the text (i.e.,
modify punctuation, or change the order of tokens). Techniques in the 
second group do not modify the vocabulary, but have a significant
impact on the number of tokens in a given mention. These augmentations
can theoretically be useful for the robustness of MD extraction models,
but only have a moderate impact in our experiments, using the PET 
dataset. We count \emph{Random Insertion}, \emph{Filler Word Augmentation},
but also \emph{Random Word Deletion} towards this group, see Figure
\ref{fig:augmentation-example} for an example taken from the augmented 
data. The final group of techniques increases the size of the 
vocabulary, while keeping mention lengths roughly the same. These
techniques are paraphrasing, aimed at preserving semantics
and the structure of textual data. Techniques using WordNet to insert
or substitute synonyms (\emph{B.100}, \emph{B.101}, 
as well as back translation methods (\emph{B.62}, \emph{B.26}) fall 
in this group. Figure~\ref{fig:augmentation-example} shows a sentence
that is augmented with the back translation method \textit{B.26}.

Figure~\ref{fig:directionlity} shows the changes in directionality 
certain data augmentation techniques have. Most techniques preserve
the direction of relation examples in the data, with the exception of
techniques \emph{B.88} (Sentence Reordering) and \emph{B.24} 
(Concatenate Random Sentences). Based on our experiments, this change
seems to be less useful than other augmentations. The improvement of
B.88 is among the worst ones of all techniques. In future work it 
could be interesting to investigate, if selectively augmenting only
certain types of relations can be helpful. Also, having a more
diverse test set, i.e., texts from different sources, like employee
notes, handbooks, and interview notes, might change the usefulness
of directionality changing data augmentation techniques.

\section{Conclusion and Future Work}\label{sec:conclusion}

In this paper we evaluated established data augmentation techniques
for use in the MD and RE steps of extracting process relevant information
from natural language texts for use in the automated generation of
business process models. To this end we selected a total of 19
distinct methods from related work, which are suitable for the given
data.

We discuss several characteristics these selected data augmentation
techniques change in the original data and how they relate to 
differences in usefulness of certain techniques for either the MD 
or RE task. We found that many of them are
useful for improving the accuracy of the current state of the art
machine learning models on the PET dataset for automated business
process model generation from natural language text. For the RE model
the $F_1$ score could be improved by up to 4.5 percentage points, the
MD model was improved by up to 2.9 percentage points. Our findings
enable researchers in the field of process model generation from 
natural language text to make more efficient use of the limited data 
available to them, enabling more precise and robust machine learning
methods for extracting business process relevant information.

In future work, we want to analyze how targeted data augmentation can 
be used to improve extraction of certain types of mentions or relations,
tackling the problem of data imbalance. Additionally we want to explore
adaptive data augmentation, where samples are selected for augmentation
by their value for model training, e.g., measured by the number of
wrong predictions.

\bibliographystyle{splncs04}
\bibliography{bibliography}

\end{document}